\documentclass[letterpaper, 10 pt, conference]{ieeeconf}
\IEEEoverridecommandlockouts
\usepackage{cite}
\usepackage{amsmath,amssymb,amsfonts}
\usepackage{algorithmic}
\usepackage{algorithm}
\usepackage{graphicx}
\usepackage{textcomp}
\usepackage{xcolor}
\usepackage[hidelinks]{hyperref}
\usepackage[caption=false, font=footnotesize]{subfig}
\usepackage{hhline}

\usepackage{tikz}
\usepackage{lipsum}

\newcommand\copyrighttext{%
  \footnotesize © 2020 IEEE. Personal use of this material is permitted. Permission from IEEE must be obtained for all other uses, in any current or future media, including reprinting/republishing this material for advertising or promotional purposes, creating new collective works, for resale or redistribution to servers or lists, or reuse of any copyrighted component of this work in other works.}
\newcommand\copyrightnotice{%
\begin{tikzpicture}[remember picture,overlay]
\node[anchor=south,yshift=10pt] at (current page.south) {\fbox{\parbox{\dimexpr\textwidth-\fboxsep-\fboxrule\relax}{\copyrighttext}}};
\end{tikzpicture}%
}

\title{\LARGE \bf
Autonomous Intruder Detection Using a ROS-Based Multi-Robot System Equipped with 2D-LiDAR Sensors
}

\author{Mashnoon Islam$^{1}$, Touhid Ahmed$^{1}$, Abu Tammam Bin Nuruddin$^{1}$, Mashuda Islam$^{1}$, Shahnewaz Siddique$^{2}$\\
{\tt\small mashnoon.islam@northsouth.edu}, %
{\tt\small touhid.ahmed@northsouth.edu}, %
{\tt\small tammam.abu@northsouth.edu},\\
{\tt\small mashuda.islam@northsouth.edu}, %
{\tt\small shahnewaz.siddique@northsouth.edu}%
\thanks{*This research has been funded by the Conference Travel and Research Grants Committee (CTRGC), North South University, Dhaka, Bangladesh.}%
\thanks{$^{1}$Research Assistant, Department of Electrical \& Computer Engineering, North South University, Dhaka, Bangladesh.}%
\thanks{$^{2}$Assistant Professor, Department of Electrical \& Computer Engineering, North South University, Dhaka, Bangladesh. IEEE Member.}%
}

\begin{document}
\maketitle
\thispagestyle{empty}
\pagestyle{empty}
\copyrightnotice

\begin{abstract}
The application of autonomous mobile robots in robotic security platforms is becoming a promising field of innovation due to their adaptive capability of responding to potential disturbances perceived through a wide range of sensors. Researchers have proposed systems that either focus on utilizing a single mobile robot or a system of cooperative multiple robots. However, very few of the proposed works, particularly in the field of multi-robot systems, are completely dependent on LiDAR sensors for achieving various tasks. This is essential when other sensors on a robot fail to  provide peak performance in particular conditions, such as a camera operating in the absence of light. This paper proposes a multi-robot system that is developed using ROS (Robot Operating System) for intruder detection in a single-range-sensor-per-robot scenario with centralized processing of detections from all robots by our central bot MIDNet (Multiple Intruder Detection Network). This work is aimed at providing an autonomous multi-robot security solution for a warehouse in the absence of human personnel.
\end{abstract}

\section{Introduction}\label{introduction}
The applications of multi-robot systems in various domains have increased dramatically in the past few decades. From automating logistics in a factory warehouse to being deployed for search and rescue missions in a hostile environment, multi-robot systems have emerged as a promising area for innovation. To that regard, the applications of multi-robot systems in the field of security, such as asset protection, patrolling, intruder detection are being extensively explored \cite{carder2018} and the recent integration of ROS in this context has opened up opportunities for devising efficient algorithms in providing autonomous robot security. The conventional robotic security system integrates different types of sensors such as LiDAR for performing Simultaneous Localization And Mapping (SLAM)\@ \citeleft \citen{cong2020}\citemid \citen{lenac2017}\citeright\@ and obstacle detection\@ \citeleft \citen{catapang2016}\citemid \citen{peng2015}\citeright\@, and camera for visual SLAM\@ \citeleft \citen{milford2008}\citemid \citen{milford2011}\citeright\@ and intruder detection\@ \citeleft \citen{chia2009}\citemid \citen{liu2005}\citeright\@. A combination of both LiDAR and camera has also been used to perform SLAM\@ \citeleft \citen{pierzchala2018}\citemid \citen{shi2019}\citeright\@ and obstacle/object detection (such as people and vehicles)\@ \citeleft \citen{asvadi2017}\citemid \citen{roth2019}\citemid \citen{guerrero2019}\citeright\@. In \cite{sankar2019}, Sankar and Tsai designed a ROS based remotely controlled mobile robot with a mounted camera for providing security in a densely crowded environment through human tracking and detection. However, the system is not autonomous and needs constant human supervision. This issue is addressed in \cite{dharmasena2019}, where, Dharmasena and Abeygunawardhana designed a ROS based indoor surveillance robot that can navigate in a known map using 2D-LiDAR while autonomously searching for an intruder through camera stream using a face recognition system developed using the OpenCV library. 

Using a single mobile robot for security application constrains the performance factor in case of greater area coverage, ultimately leading to longer time of convergence to the intruder, potentially resulting in the escape of the intruder. In contrast, the implementation of a multi-robot system\@ \citeleft \citen{khamis2015}\citemid \citen{karapetyan2017}\citeright\@ mitigates these problems and provides better security. To that regard, there has been quite some work in this sector which highlights the capability of detecting intruders in a multi-robot system. We present three particular cases in that respect. In \cite{folgado2007}, Folgado et al. have designed and simulated a multi-robot surveillance system in a pre-defined wide enclosed area where an intruder is detected based on motion detection from video frames captured by the robots' cameras. The system requires the security robots to be stationary while searching for a single intruder in the pre-defined map. Based on the location of the detection, the robots navigate to the intruder for apprehension. In addition, the system assumes that a detection is only valid if something around the security robot moves while it is in stationary state during the set interval. The assumption itself draws a major flaw on this system as the intruder can virtually be invisible to the security robots by standing still during the set interval. This trick could be essential for the escape of the intruder. In \cite{trigui2012}, Trigui et al. proposed a multi-robot surveillance system that uses a distributed wireless sensor network with the aim of solving the coordination problem among multiple robots during indoor surveillance. Among the three co-ordination strategies presented, the centralized approach has shown better results for the time of convergence to the intruder. However, the system becomes susceptible in case of sensor failure. In addition, since the mobile robots, do not have the global knowledge about the environment, it is possible for multiple robots to patrol a certain part of the environment, leaving the rest unguarded. In \cite{tuna2012}, Tuna et al. have proposed a system architecture for an autonomous intruder detection system through multi-sensor fusion using wireless networked robots. The work has focused on the following aspects of the system: coordination and task allocation, communication and map-based intruder detection. However, the work has only highlighted on reducing the positional errors of the robots using multi-sensor fusion. Hence, the work is inconclusive over the intruder detection system itself. 

To precisely detect intruders, most of the above systems have used cameras in conjunction with LiDAR sensors. However, the detection mechanism can fail severely in the event of camera hardware failure or poor lighting conditions. Although the LiDAR hardware can also fail, it performs well in poor lighting conditions. To the best of the authors’ knowledge, there has not been any work that uses 2D-LiDAR for intruder detection in the context of both single and multi-robot systems. Hence, this is the first work that proposes a centralized and scalable system that deploys a team of robots to detect intruders in a priori map using 2D-LiDAR.

Further sections of this paper have been arranged in the following order: Section \ref{system_overview} provides an overview of the system and the software components that have been used. Section \ref{simulation_layout} discusses the simulation testbeds setup for the proposed system. Section \ref{intruder_detection_pipeline} illustrates the implementation of MIDNet with relevant algorithms, diagrams and tables. Section \ref{experiments_and_results} provides the experimental data that have been recorded while conducting the test trials with the system prototype and validates the system performance using specific evaluation metrics. Section \ref{conclusion} concludes the paper by providing a general insight into the proposed system.

\section{System Overview}\label{system_overview}
\subsection{Map Construction}\label{map_construction}
The proposed system for detecting unwanted entities in a pre-defined map consists of initially building the map itself. This map is an occupancy grid map generated by the ROS package \emph{gmapping}, which is a wrapper that runs the \emph{OpenSLAM GMapping} algorithm of Grisetti et al.\@ \citeleft \citen{grisetti2005}\citemid \citen{grisetti2007}\citeright\@ in the ROS ecosystem. To improve our robots' pose estimates, we have used the ROS package \emph{robot\_localization} of Moore et al. \cite{moore2016} to run the Unscented Kalman Filter algorithm proposed by Julier et al. \cite{julier1997}. For autonomous exploration and map merging, we have used the ROS packages \emph{explore\_lite} and \emph{multirobot\_map\_merge}, both of which have been developed by Hörner\@ \cite{horner2016}. Depending on the quality of the final maps that have been produced by single-robot exploration and multi-robot exploration, the best map is saved using the ROS package \emph{map\_server} for further processing in the next steps. Each robot in our system uses the ROS package \emph{move\_base}, which packs an array of additional ROS packages that are used for navigation. \emph{navfn}, a ROS package that is included in \emph{move\_base}, uses Dijkstra's algorithm \cite{dijkstra1959} to find an optimal global path through a global cost map produced by \emph{costmap\_2d}, which is another ROS sub-package of \emph{move\_base} that runs an implementation of the layered grid map approach proposed by Lu et al.\@ \cite{lu2014}. The package \emph{base\_local\_planner} includes an implementation of a local path planner for obstacle avoidance, which also uses \emph{costmap\_2d}, the Dynamic Window approach \cite{fox1997} and the Trajectory Rollout approach \citeleft \citen{kelly1994}\citemid \citen{gerkey2008}\citeright. Lastly, a separate instance of \emph{costmap\_2d} is used, with parameters different from that of the \emph{costmap\_2d} instances of \emph{move\_base}, to build a global costmap that is essential for our system. The best costmap is converted into a binary grid map using a specified threshold cost value, which ranges from 0 to 100, and is saved to disk using the \emph{map\_server} ROS package.

\subsection{Intruder Detection}\label{intruder_detection}
This part of the system uses the same \emph{move\_base} package mentioned in Section \ref{map_construction} for navigation purposes of each robot. A separate instance of \emph{costmap\_2d} is spawned again for each robot with parameters identical to that of the instance of \emph{costmap\_2d} for the global costmap in the previous subsection - this time for creating a local grid map that behaves like the ``eyes" of the robot (this is further explained in Section \ref{thresholding_incoming_grid_maps}). The saved map from Section \ref{map_construction} is served to this part of the system using the \emph{map\_server} ROS package. Similar to Section \ref{map_construction}, \emph{robot\_localization} has also been used in this phase to improve positional accuracy of the robots. Finally, the \emph{midnet} ROS node, which runs MIDNet in the ROS ecosystem, uses information from all of our robot instances to detect intruders and determine their approximate locations in the priori map.

\section{Simulation Layout}\label{simulation_layout}
We have created two simulation testbeds in Gazebo to analyze the performance of our proposed system. We have chosen Gazebo as it supports real-world physics (using the Open Dynamics Engine) and sensor noise. The first simulation testbed, which we refer to as Map 1 (shown in Fig 1(a)) is an $8m \times 8m$ ground plane with a $6m \times 6m$ wooden square enclosure on it. This testbed has been set up to measure the performance of MIDNet in classifying and localizing intruders. Both stationary intruders and intruders in motion are introduced during the testing phase. With that said, we also have to make a data-driven conclusion on how effective it is to converge to the intruders if the number of security robots is increased. To conduct this analysis, we have created a second testbed, Map 2 (shown in Fig 1(b)), which is a $24m \times 24m$ ground plane with a $20m \times 20m$ bounded region. The wooden boundary has two door-like openings on opposite sides of the region to allow any moving entity to enter or exit the bounded region. Unlike Map 1, only mobile intruders are introduced in Map 2. Lastly, all mobile robots that are used in both Map 1 and Map 2 are Turtlebot3 Burgers. 

\begin{figure}
	\centering
	\subfloat[Map 1: $8m \times 8m$ ground  plane  with $6m \times 6m$ enclosure. This is used to measure MIDNet's detection performance.]{%
		\includegraphics[width=0.82\linewidth]{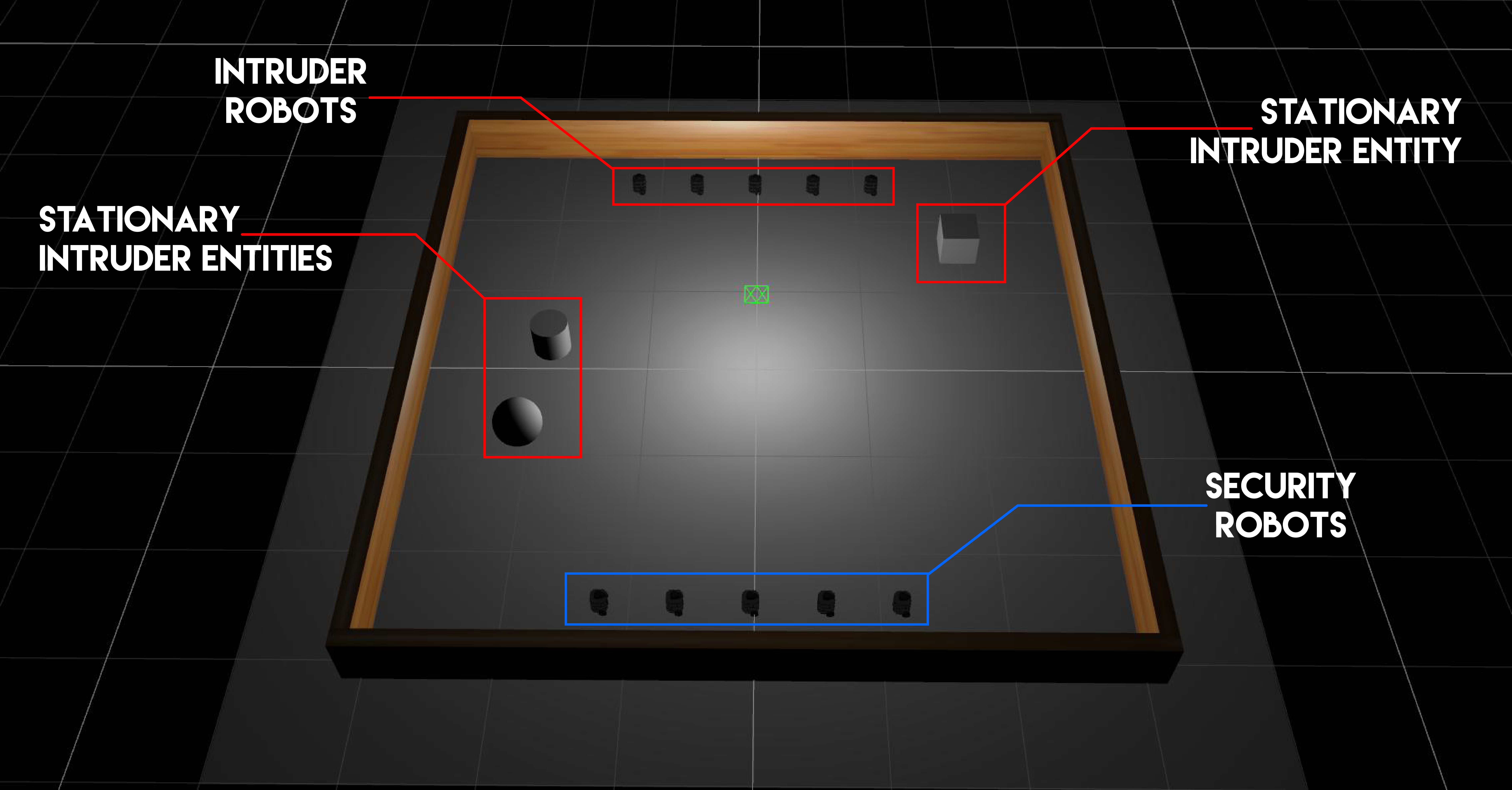}}
    \hfill
	\subfloat[Map 2: $24m \times 24m$ ground plane with $20m \times 20m$ bounded region. This is used to make a data-driven conclusion about the effectiveness of increasing security robots.]{%
		\includegraphics[width=1.0\linewidth]{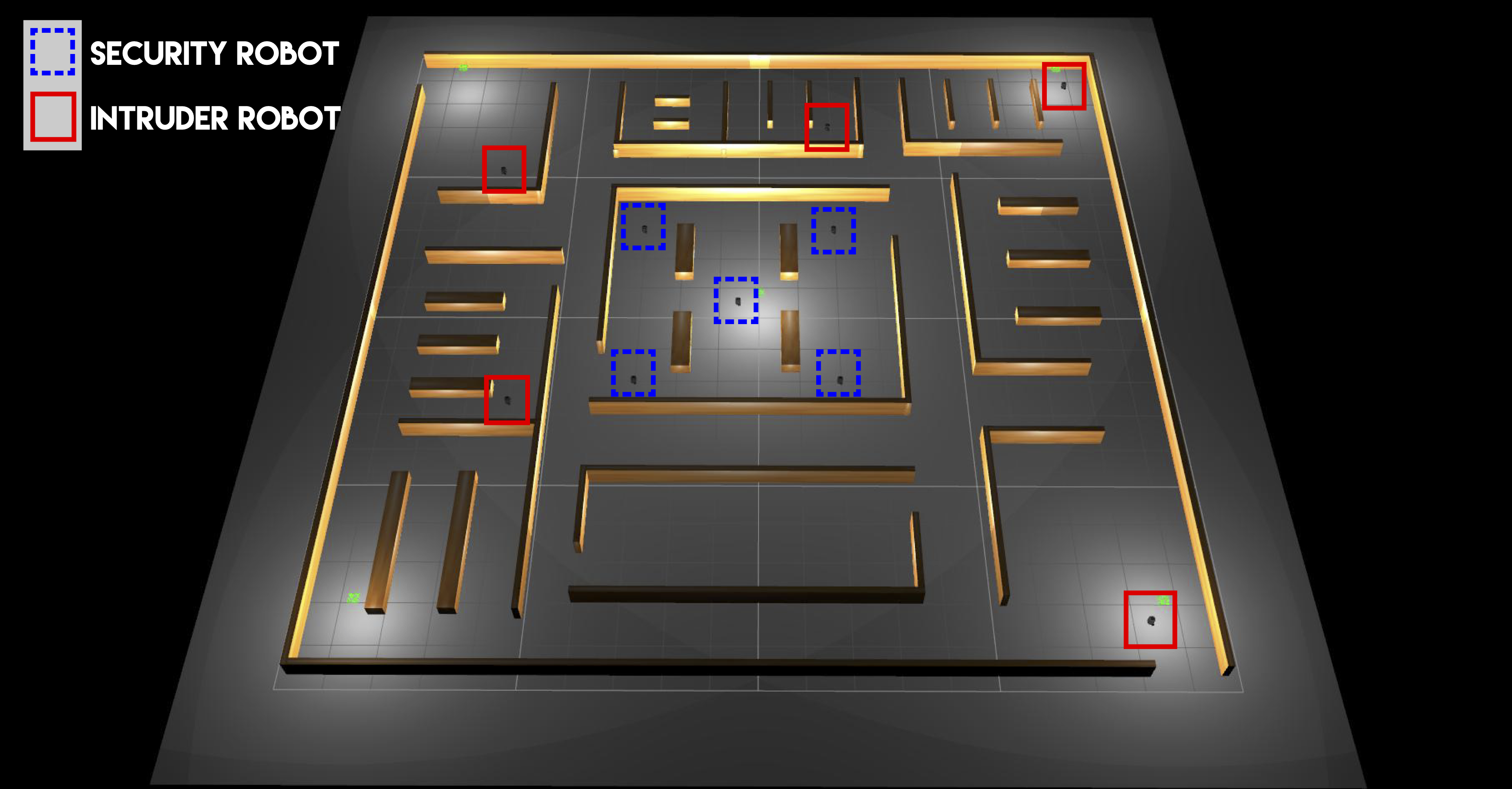}}
  \caption{Simulation Testbeds}
  \label{simulation_testbeds} 
\end{figure}

\section{Intruder Detection Pipeline}\label{intruder_detection_pipeline}
\subsection{Conversion of Incoming Grid Maps to Binary Grid Maps}\label{thresholding_incoming_grid_maps}
The process of our intruder detection pipeline starts with converting the local costmaps of each robot that is generated by \emph{costmap\_2d} into black-and-white grid maps. The equation that is associated with this process is given in Eq. (\ref{grid_map_conversion}).

\begin{equation}\label{grid_map_conversion}
	E_{i, j} = 
	\begin{cases}
		255, & \mathit{if\ } 0 \leq x < \mathit{thresh}\\
		0, & \mathit{if\ } \mathit{thresh} \leq x \leq 100\\
	\end{cases}
\end{equation}
Here, $E_{i, j}$ is the element of the grid map $E$ that is located at the position $(i, j)$. The costmaps initially contain cells having values ranging from 0 to 100. Any given cell having a cost value greater than or equal to $\mathit{thresh}$ is assigned a color value of $0$ in grayscale, which represents an occupied grid cell. If otherwise, the cell is assigned a color value of $255$, which represents either a free or an unknown grid cell. The value of $\mathit{thresh}$ is to be tuned till the system produces little or no false positive detections (caused by LiDAR sensor noise), which is done by setting the value of $\mathit{thresh}$ higher than that of the global grid map in Section \ref{map_construction}. 

\subsection{Accumulating Information from Each Robot}\label{information_accumulation_from_each_robot}
\subsubsection{Cropping Relevant Section of Map}\label{crop_relevant_section_of_map}
To extract useful information from the surroundings of a given robot, the relevant portion is cropped out of the priori global grid map and compared to the current local grid map. The equation for calculating the left-top coordinate $({x_1}^{grid}, {y_1}^{grid})$ necessary for the crop is given in Eq. (\ref{crop_left_top_coord}). 

\begin{equation}\label{crop_left_top_coord}
	\begin{bmatrix}
		{x_1}^{\mathit{grid}}\\
		{y_1}^{\mathit{grid}}\\
	\end{bmatrix}
	= 
	\begin{bmatrix}
		{x_{c}}^{\mathit{grid}}\\
		{y_{c}}^{\mathit{grid}}\\
	\end{bmatrix}
	-
	\frac{1}{2}
	\begin{bmatrix}
		w^{\mathit{grid}}\\
		l^{\mathit{grid}}\\
	\end{bmatrix}
\end{equation}
The equation associated with the calculation of the right-bottom coordinate $({x_2}^{\mathit{grid}}, {y_2}^{\mathit{grid}})$  for the crop is given in Eq. (\ref{crop_right_bottom_coord}).

\begin{equation}\label{crop_right_bottom_coord}
	\begin{bmatrix}
		{x_2}^{\mathit{grid}}\\
		{y_2}^{\mathit{grid}}\\
	\end{bmatrix}
	= 
	\begin{bmatrix}
		{x_{c}}^{\mathit{grid}}\\
		{y_{c}}^{\mathit{grid}}\\
	\end{bmatrix}
	+
	\frac{1}{2}
	\begin{bmatrix}
		w^{\mathit{grid}}\\
		l^{\mathit{grid}}\\
	\end{bmatrix}
\end{equation}
For both Eq. (\ref{crop_left_top_coord}) and Eq. (\ref{crop_right_bottom_coord}), $w^{\mathit{grid}}$ is the width of the local grid map and $l^{\mathit{grid}}$ is the length of the local grid map, both in pixels. $({x_{c}}^{\mathit{grid}}, {y_{c}}^{\mathit{grid}})$ is the current position of the robot in pixel coordinates, which is calculated using Eq. (\ref{current_robot_position}). $({x_{c}}^{\mathit{grid}}, {y_{c}}^{\mathit{grid}})$ is always in the middle of the square or rectangular window that $({x_{1}}^{\mathit{grid}}, {y_{1}}^{\mathit{grid}})$ and $({x_{2}}^{\mathit{grid}}, {y_{2}}^{\mathit{grid}})$ create (which is in fact the local grid map itself). 

\begin{equation}\label{current_robot_position}
	\begin{bmatrix}
		{x_c}^{\mathit{grid}}\\
		{y_c}^{\mathit{grid}}\\
	\end{bmatrix}
	= 
	\frac{1}{r}
	\left(
	\begin{bmatrix}
		{x_{c}}^{\mathit{world}}\\
		{y_{c}}^{\mathit{world}}\\
	\end{bmatrix}
	-
	\begin{bmatrix}
		{x_{o}}^{\mathit{world}}\\
		{y_{o}}^{\mathit{world}}\\
	\end{bmatrix}
	\right)
\end{equation}
Here, $({x_{c}}^{\mathit{world}}, {y_{c}}^{\mathit{world}})$ is the current position of the robot in the real world, measured in metres ($m$), $({x_{o}}^{\mathit{world}}, {y_{o}}^{\mathit{world}})$ is the origin of the priori global grid map in metres and $r$ is the resolution of the global grid map, in metres per occupancy grid cell.

\begin{figure}
	\centering
	\includegraphics[width=0.99\linewidth]{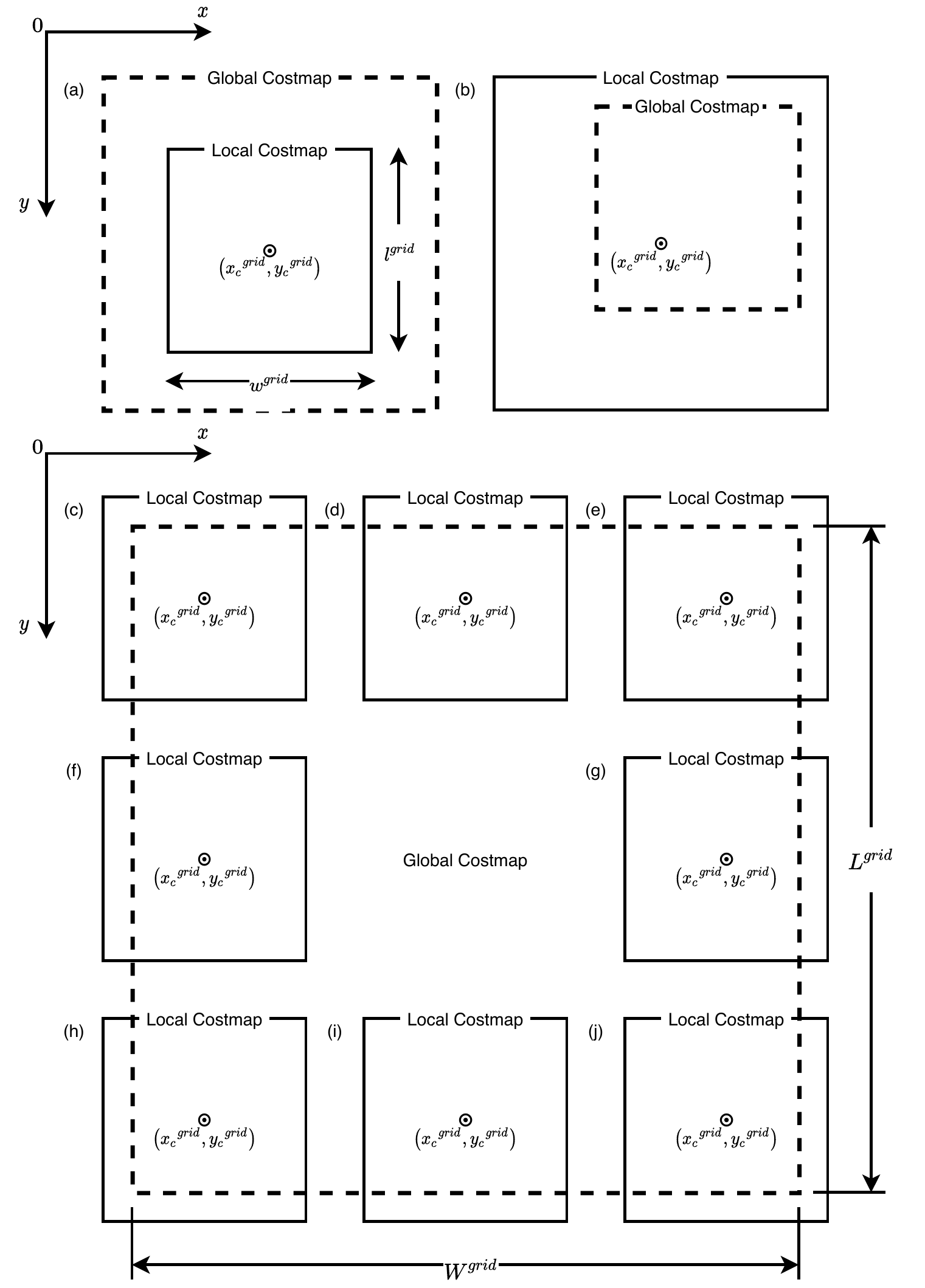}
	\caption{Special cases for cropping. All local grid maps in this figure have dimensions $(l^{\mathit{grid}} \times w^{\mathit{grid}})$, while all global grid maps have dimensions $(L^{\mathit{grid}} \times W^{\mathit{grid}})$. Relevant to Section \ref{crop_relevant_section_of_map}.}
	\label{special_cases_for_cropping}
\end{figure}

The maximum width and length of the local grid map depend on the maximum usable range of the LiDAR sensor being used, and they are usually smaller than that of the priori global grid map, as shown in Fig. \ref{special_cases_for_cropping}(a). This means that the values of $({x_1}^{\mathit{grid}}, {y_1}^{\mathit{grid}})$ and $({x_2}^{\mathit{grid}}, {y_2}^{\mathit{grid}})$ will not be outside the priori global map coordinate system. This may not always be the case. In one scenario, the local grid map may be larger than the global grid map, as shown in Fig. \ref{special_cases_for_cropping}(b). In another scenario, when the robot goes near the boundary of the priori global grid map, it is highly possible that ${x_1}^{\mathit{grid}}$, ${y_1}^{\mathit{grid}}$, ${x_2}^{\mathit{grid}}$ and ${y_2}^{\mathit{grid}}$ may be outside the global map's boundary, as shown in Fig. \ref{special_cases_for_cropping}(c)-(j). Algorithm \ref{find_cropping_coordinates} solves this issue by calculating two new coordinates $({x_1}^{\mathit{lmap}}, {y_1}^{\mathit{lmap}})$ and $({x_2}^{\mathit{lmap}}, {y_2}^{\mathit{lmap}})$ for the local grid map and by calculating two other coordinates $({x_1}^{\mathit{gmap}}, {y_1}^{\mathit{gmap}})$ and $({x_2}^{\mathit{gmap}}, {y_2}^{\mathit{gmap}})$ for the global grid map. This ensures that for the special cases in Fig. \ref{special_cases_for_cropping}, the new coordinates that are going to be used for cropping lie inside the boundary of the global grid map. Consequently, the cropped sections of the two maps have equal length and width. 

\begin{algorithm}\caption{Find Cropping Coordinates}\label{find_cropping_coordinates}
\begin{algorithmic}
	\STATE \COMMENT{related to Fig. \ref{special_cases_for_cropping}(b, c, f, h)}
	\IF{${x_1}^{\mathit{grid}} < 0$}
		\STATE ${x_1}^{\mathit{lmap}} \leftarrow \left \lvert {x_1}^{\mathit{grid}} \right \rvert$
		\STATE ${x_1}^{\mathit{gmap}} \leftarrow 0$
	\ENDIF
	\STATE \COMMENT{related to Fig. \ref{special_cases_for_cropping}(b, c, d, e)}
	\IF{${y_1}^{\mathit{grid}} < 0$}
		\STATE ${y_1}^{\mathit{lmap}} \leftarrow  \left \lvert {y_1}^{\mathit{grid}} \right \rvert$
		\STATE ${y_1}^{\mathit{gmap}} \leftarrow 0$
	\ENDIF
	\STATE \COMMENT{related to Fig. \ref{special_cases_for_cropping}(b, e, g, j)}
	\IF{${x_2}^{\mathit{grid}} > W^{\mathit{grid}}$}
		\STATE ${x_2}^{\mathit{lmap}} \leftarrow w^{\mathit{grid}} - ({x_2}^{\mathit{grid}} - W^{\mathit{grid}})$
		\STATE ${x_2}^{\mathit{gmap}} \leftarrow W^{\mathit{grid}}$
	\ENDIF
	\STATE \COMMENT{related to Fig. \ref{special_cases_for_cropping}(b, h, i, j)}
	\IF{${y_2}^{\mathit{grid}} > H^{\mathit{grid}}$}
		\STATE ${y_2}^{\mathit{lmap}} \leftarrow h^{\mathit{grid}} - ({y_2}^{\mathit{grid}} - H^{\mathit{grid}})$
		\STATE ${y_2}^{\mathit{gmap}} \leftarrow H^{\mathit{grid}}$
	\ENDIF
\end{algorithmic}
\end{algorithm}

\subsubsection{Processing Cropped Section of the Grid Map}\label{processing_cropped_section_of_map}
\begin{figure}
	\centering
	\includegraphics[width=0.6\linewidth]{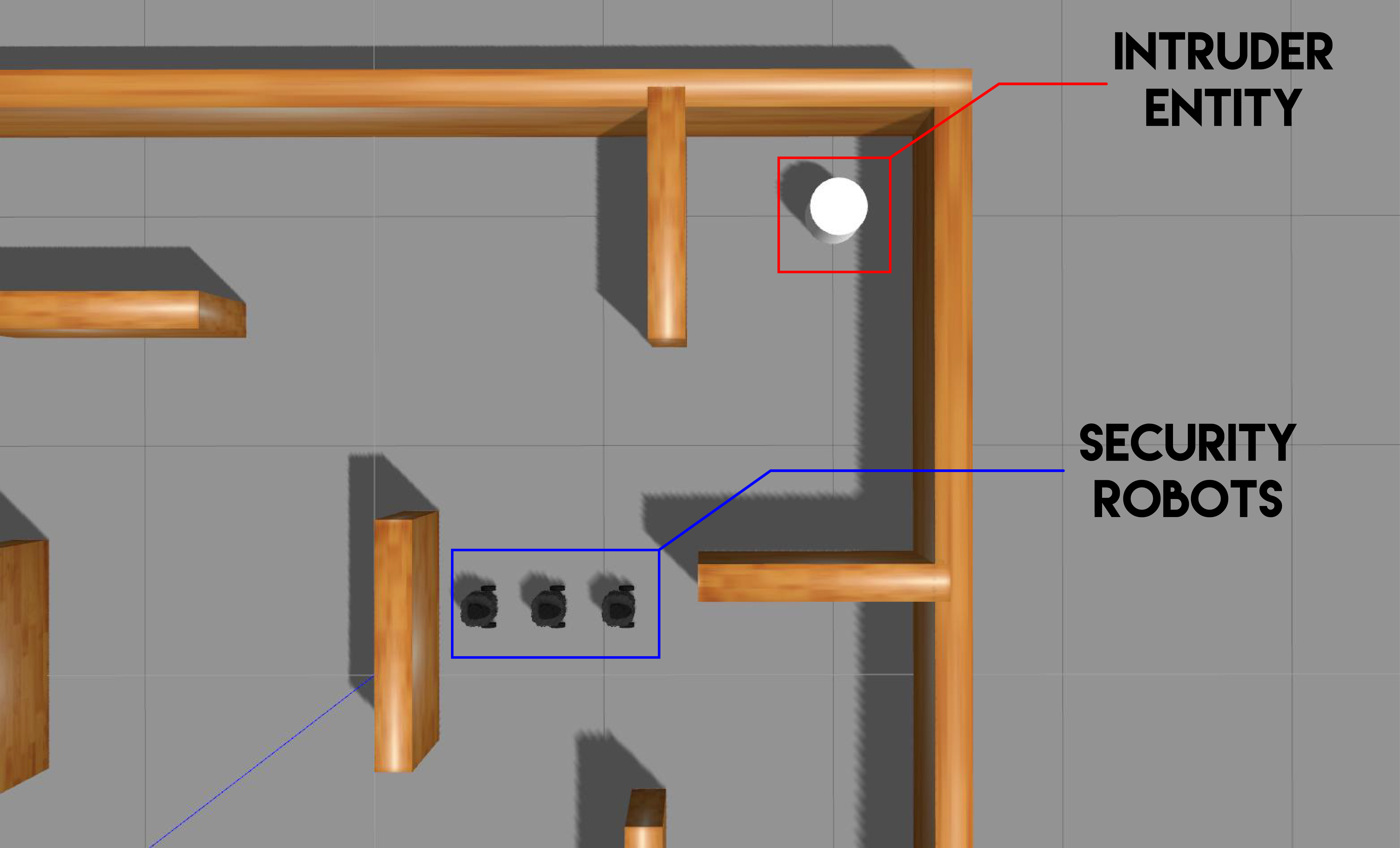}
	\caption{Aerial view of part of the real world relevant to Section \ref{information_accumulation_from_each_robot}.
	}
	\label{real_world_scanned}
\end{figure}

\begin{figure} 
	\centering
	\subfloat[Local grid maps $A$, $B$, $C$ and $D$ for the first robot.]{%
       \includegraphics[width=1.0\linewidth]{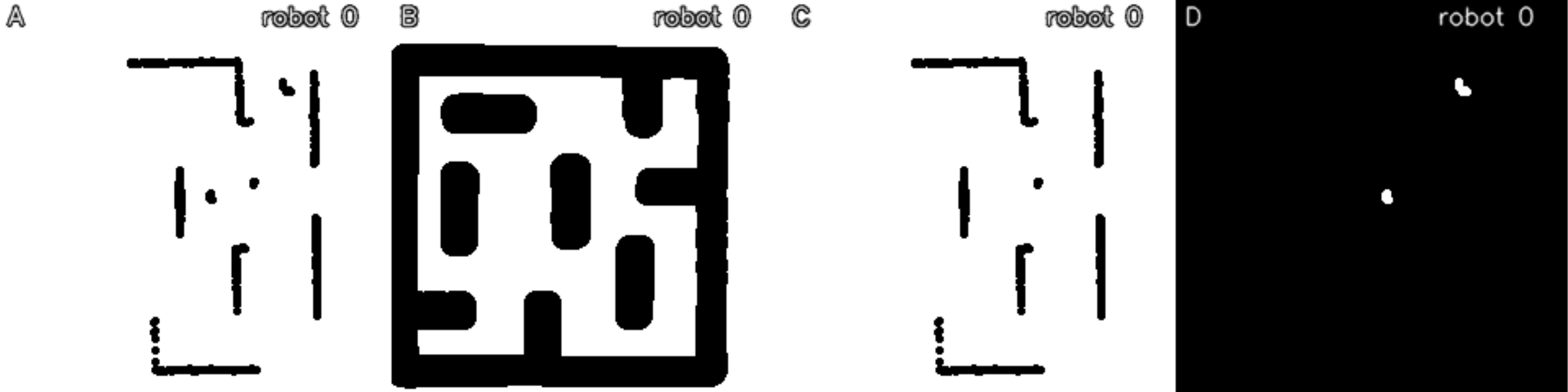}}
    \\
	\subfloat[Local grid maps $A$, $B$, $C$ and $D$ for the second robot.]{%
        \includegraphics[width=1.0\linewidth]{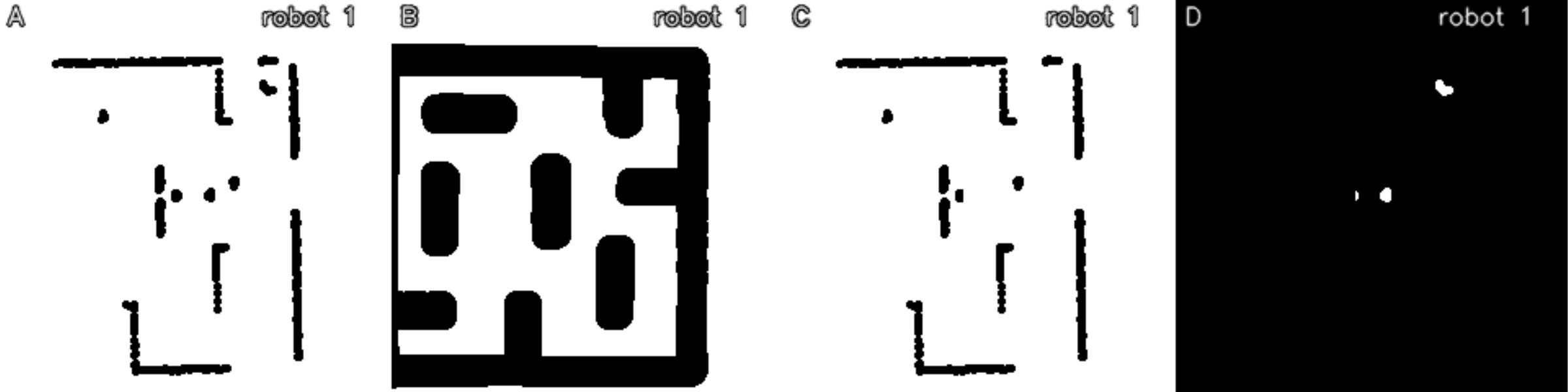}}
    \\
	\subfloat[Local grid maps $A$, $B$, $C$ and $D$ for the third robot.]{%
        \includegraphics[width=1.0\linewidth]{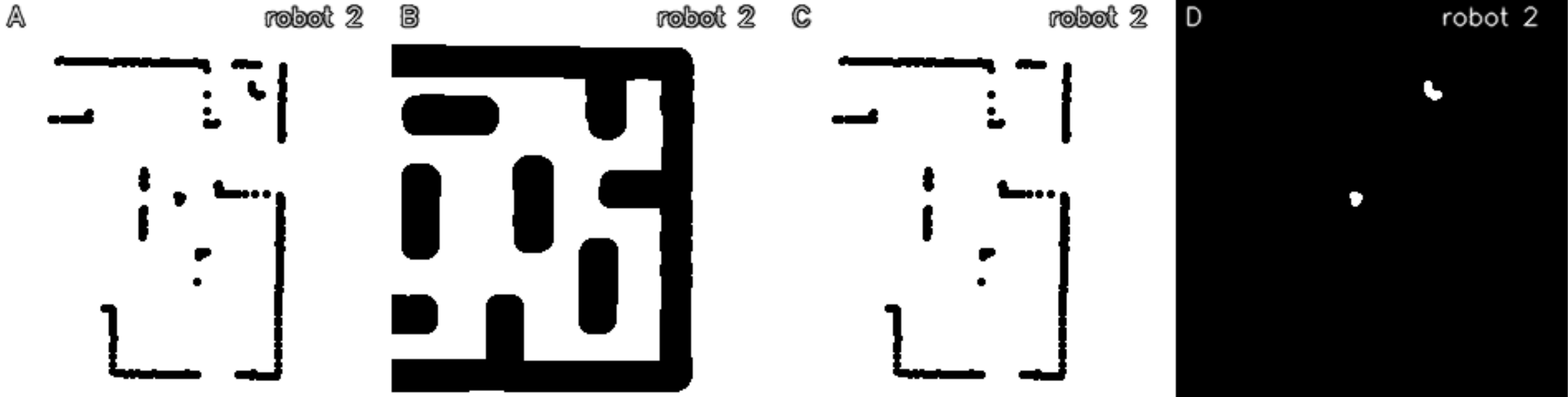}}
  \caption{Processed local grid maps relevant to Section \ref{information_accumulation_from_each_robot}.}
  \label{local_grid_maps_processed} 
\end{figure}

The portion of the real-world surroundings that is relevant to the rest of this section is given in Fig. \ref{real_world_scanned}. After the process in Section \ref{crop_relevant_section_of_map}, we have a local grid map $A$ (which may be cropped) and a cropped section of the priori global grid map $B$ with equal length $l_{\mathit{new}}$ and equal width $w_{\mathit{new}}$. The goal is to compare $A$ and $B$ such that, every grid cell that has a value of $0$ in both $A$ and $B$ are filtered out in a new grid map $C$. This involves an elementwise $\mathit{OR}$ operation between $A$ and $B$, as shown in Eq. (\ref{bitwise_or}). This step is necessary for two reasons, one is that the grid map with real-time data $A$ will not include features of regions that are masked by obstacles. The other reason is that the cropped priori global grid map $B$ has more occupied cells than the local grid map $A$, which is the result of the global grid map being slightly more inflated than the local grid map. Thus, $C$ visually appears to have the common occupancy features of both $A$ and $B$, which is exactly what we need. 

\begin{equation}\label{bitwise_or}
	\underset{(w_{\mathit{new}} \times l_{\mathit{new}})}{C} = \underset{(w_{\mathit{new}} \times l_{\mathit{new}})}{A}  \vee \underset{(w_{\mathit{new}} \times l_{\mathit{new}})}{B}
\end{equation}
Now, the local grid map $A$ has live data of the robot's surroundings, which also includes occupancy information of potential intruders. To filter out the intruder occupancy grid cells, an element-wise subtraction between $C$ and $A$, followed by an elementwise calculation of absolute values, is performed, as shown in Eq. (\ref{absdiff}). In the field of Computer Vision, this is known as Background Subtraction. This process results in a new grid map $D$ in which cells that are not occupied in $C$ (and in turn $B$) but occupied in $A$ have a value of $255$, and the rest of the cells carry a value of $0$. The cells with a value of $255$ in $D$ are therefore the cells occupied by the intruder(s) in the local grid map $A$. The grid maps $A$, $B$, $C$ and $D$ for all the three robots in our system are displayed in Fig. \ref{local_grid_maps_processed}.

\begin{equation}\label{absdiff}
	\underset{(w_{\mathit{new}} \times l_{\mathit{new}})}{D} = \left \lvert \underset{(w_{\mathit{new}} \times l_{\mathit{new}})}{A}  - \underset{(w_{\mathit{new}} \times l_{\mathit{new}})}{C} \right \rvert 
\end{equation}

\subsubsection{Detecting Contours in Cropped Section of the Grid Map}\label{detecting_contours_in_cropped_section_of_map}
The white sections in $D$ are the parts where the real-time local grid map $A$ does not match with the cropped priori local grid map $B$. The OpenCV function \emph{findContours} has been used to obtain these sections as list of contours. The implementation of \emph{findContours} is based on the work of Suzuki and Abe \cite{suzuki1985}.

\subsubsection{Determining Bounding Boxes of Contours Detected}\label{determining_bounding_boxes_of_contours_detected}
\begin{figure} 
	\centering
	\subfloat[]{%
		\includegraphics[width=0.33\linewidth]{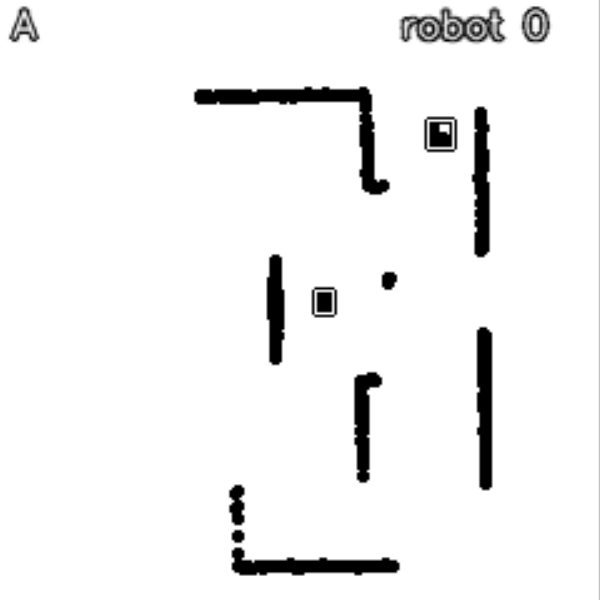}}
	\subfloat[]{%
		\includegraphics[width=0.33\linewidth]{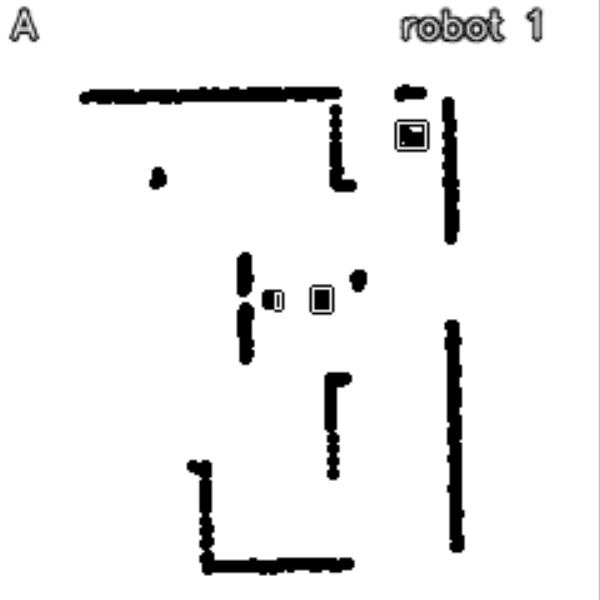}}
	\subfloat[]{%
        \includegraphics[width=0.33\linewidth]{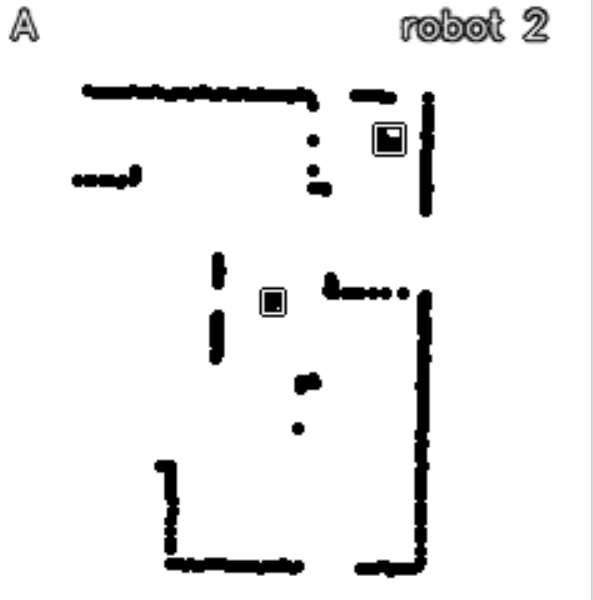}}
  \caption{Local grid maps $A$ with detection boxes. Relevant to Section \ref{information_accumulation_from_each_robot}.}
  \label{local_grid maps_with_detections} 
\end{figure}

To draw an up-right bounding rectangle for a given contour, the minimum and maximum values of each dimension of the contour points have to be figured out, which is exactly what the OpenCV function \emph{boundingRect} does. Fig. \ref{local_grid maps_with_detections} shows the grid map $A$ for each robot after the bounding boxes have been drawn. 

\subsection{Centralized Information Processing by MIDNet}\label{centralized_information_processing}

\subsubsection{Removing Ally Robot Detections From List of Detections}\label{filtering_ally_robot_detections_from_list_of_detections}
Before fusing information about intruders from all the robots, it has to be made sure that each robot does not identify its allies as intruders. In Fig. \ref{local_grid maps_with_detections}, it can be seen that each robot is in fact detecting the ally visible to it as a potential intruder, further confirmed after comparing with the real-world aerial view in Fig. \ref{real_world_scanned}. This problem is solved by our central instance MIDNet. Algorithm \ref{remove_ally_detections} describes the steps on how MIDNet removes the detections that are related to the ally robots.

\begin{figure}
	\centering
	\includegraphics[width=0.6\linewidth]{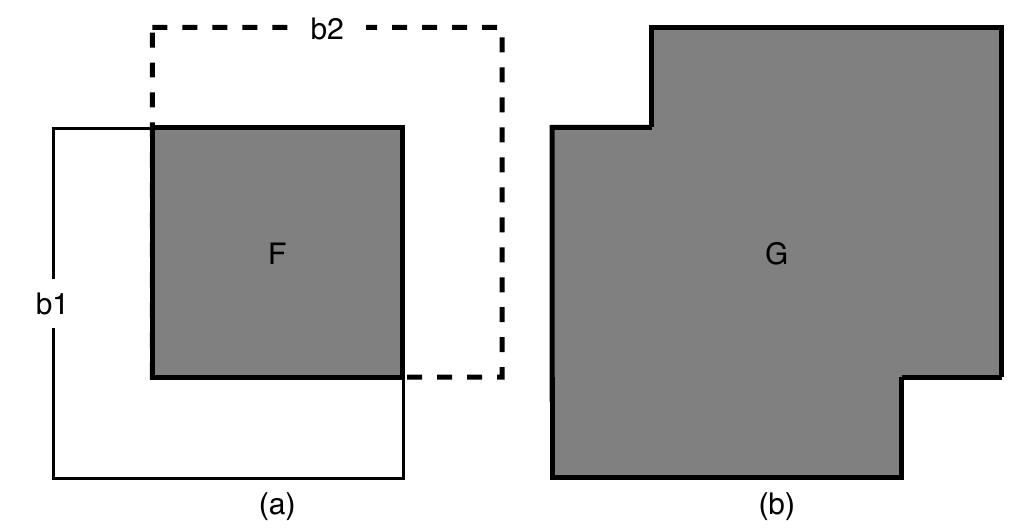}
	\caption{Visual demonstration of Intersection over Union (IoU), relevant to Section \ref{centralized_information_processing}.}
	\label{iou_demonstration}
\end{figure}

\begin{algorithm}\caption{Remove Ally Detections}\label{remove_ally_detections}
\textbf{global variable:} \text{a list of robot instances $R$ of size $n$}\\
\{$\mathit{di}$: diameter of robot, $t$: IoU score threshold, \textbf{bbox}: returns square bounding box, $d$: list of detections, \textbf{iou}: returns IoU score, defined in Eq. (\ref{iou_calculation})\}\\
\textbf{function rad($\mathit{di}$, $t$):}
\begin{algorithmic}
	\FOR{$i$ in $R$}
		\STATE $O \leftarrow R$
		\STATE $O$.\textbf{pop($i$)}
		\STATE $b \leftarrow$ \textbf{bbox($\mathit{i.pos}$, $\mathit{di}$)} 
		\FOR{$j$ in $O$}
			\FOR{$k$ in $\mathit{j.d}$}
				\STATE $s \leftarrow $ \textbf{iou($b$, $k$)}
				\IF{$s \geq t$}
					\STATE $j$.\textbf{pop($k$)}
					\STATE \textbf{break}
				\ENDIF
			\ENDFOR
		\ENDFOR
	\ENDFOR
\end{algorithmic}
\textbf{end function}
\end{algorithm}
Here, in Algorithm \ref{remove_ally_detections}, $\mathit{di}$ is the diameter of the circle that the shape of each robot is inscribed in. This shape is the pattern seen from an aerial viewpoint. The function \emph{bbox} returns the coordinates needed to draw a square bounding box representing the area covered by the ally in that ally's current position. $d$ contains the list of detections accumulated by a given robot and belongs to a robot instance from the global list $R$. The function \emph{iou} returns the IoU score between two overlapping regions, and the mathematical expression for this score is given in Eq. (\ref{iou_calculation}). Finally, $t$ is the IoU score threshold below which two overlapping bounding boxes of interest are not considered to be bounding boxes representing the same object in the real world. Thus, if the IoU score of the ally bounding box and the bounding box of one of the detections is equal to or greater than this threshold, that particular detection is deleted from the list of detections of the robot.

\begin{equation}\label{iou_calculation}
	\mathit{iou} = \frac{F}{G} = \frac{\mathit{b1} \cap \mathit{b2}}{\mathit{b1} \cup \mathit{b2}}
\end{equation}
Here, $F$ is the area of intersection of rectangle $\mathit{b1}$ and rectangle $\mathit{b2}$ (shown in Fig. \ref{iou_demonstration}(a)) and $G$ is the union area of $\mathit{b1}$ and $\mathit{b2}$, as shown in Fig. \ref{iou_demonstration}(b).

\subsubsection{Merging Overlapping Detections From List of Detections by MIDNet}\label{merging_overlapping_detections_from_list_of_detections}
The ally detections have been filtered out, and the only detections that are left must be related to the intruders. Detections from all robots could be drawn into the deflated version of the priori global map and the system pipeline could have ended in this stage. However, while drawing the detections, an issue of duplicate detections, and thus, overlapping bounding boxes are seen. This happens when more than one robot detects the same intruder, which can be observed in Fig. \ref{local_grid maps_with_detections}, where all the three robots can see the same intruder object at the top corner of the real world. While this does not cause loss of information about intruders or their locations, it gives an overestimation of the number of intruders in the real world. It also clutters the priori global map where the bounding boxes are drawn to display intruder information and metadata. To solve this problem, Algorithms \ref{divide_and_conquer} and \ref{fuse_detections} have been developed to overcome the duplicate detection issue.

\begin{algorithm}\caption{Divide and Conquer}\label{divide_and_conquer}
\textbf{global variable:} \text{a list of robot instances $R$ of size $n$}\\
\{$l$: left index, $r$: right index, $d$: list of detections, $t$: IoU score threshold, \textbf{fd}: returns fused list of detections, given in Algorithm \ref{fuse_detections}\}\\
\textbf{function doc($l$, $r$, $t$):}
\begin{algorithmic}
\IF{$l = r$}
\RETURN $\mathit{R[l].d}$
\ENDIF
\STATE $q \leftarrow \lfloor (l+r) / 2 \rfloor$
\STATE $L1 \leftarrow$ \textbf{doc($l$, $q$)}
\STATE $L2 \leftarrow$ \textbf{doc($q + 1$, $r$)}
\STATE $L \leftarrow$ \textbf{fd($L2$, $L1$)} 
\RETURN $L$
\end{algorithmic}
\textbf{end function}
\end{algorithm}
Algorithm \ref{divide_and_conquer} breaks down the list of robot instances $R$, of size $n$, in a recursive fashion, using the divide-and-conquer approach. The recursion can be launched by calling the function \emph{doc} in Algorithm \ref{divide_and_conquer} and passing the parameters $l$ as $1$, $r$ as $n$ and $t$ as the IoU threshold value of choice. To start the recursion, if the indexing is from $0$, the values of $l$ and $r$ decreases by $1$. The recursive process returns a fused list $L$, which does not have any overlapping bounding boxes, and thus provides better information about the intruders in the real world. 

\begin{figure}
	\centering
	\includegraphics[width=1.0\linewidth]{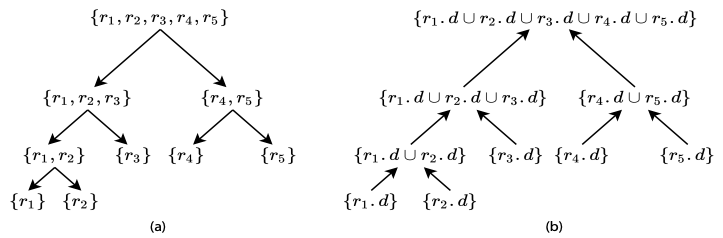}
	\caption{(a) Divide-and-conquer tree. (b) Combine tree. Relevant to Section \ref{merging_overlapping_detections_from_list_of_detections}.}
	\label{divide_and_conquer_tree}
\end{figure}

Assume that the system has a list of robot instances $R = \{r_1, r_2, r_3, r_4, r_5\}$. Each robot instance has a member $d$, which contains the list of detections it has accumulated. The divide-and-conquer tree produced by Algorithm \ref{divide_and_conquer} is given in Fig. \ref{divide_and_conquer_tree}(a).
Algorithm \ref{divide_and_conquer} uses Algorithm \ref{fuse_detections} to find a union of the two lists being combined at each step of the combination process shown in \ref{divide_and_conquer_tree}(b).
\begin{algorithm}\caption{Fuse Detections}\label{fuse_detections}
\{$\mathit{S1}$: first set/list, $\mathit{S2}$: second set/list, $t$: IoU score threshold, \textbf{iou}: return IoU score, defined in Eq. (\ref{iou_calculation})\}\\
\textbf{function fd($\mathit{S1}$, $\mathit{S2}$, $t$):}
\begin{algorithmic}
\STATE $S \leftarrow \mathit{S2}$
\FOR{$i$ in $\mathit{S1}$}
	\STATE $\mathit{ip} \leftarrow \FALSE$
	\FOR{$j$ in $\mathit{S2}$}
		\STATE $s \leftarrow $ \textbf{iou($i$, $j$)}
		\IF{$s \geq t$}
			\IF{\textbf{area($i$)} $>$ \textbf{area($j$)}}
				\STATE $S$[\textbf{index($j$)}] $\leftarrow i$
			\ENDIF
			\STATE $\mathit{ip} \leftarrow \TRUE$
			\STATE \textbf{break}
		\ENDIF
	\ENDFOR
	\IF{$\mathit{ip} = \FALSE$}
		\STATE $S \leftarrow L \cup i$
		\STATE $ip \leftarrow \TRUE$
	\ENDIF
\ENDFOR
\RETURN $S$
\end{algorithmic}
\textbf{end function}
\end{algorithm}
In Algorithm \ref{fuse_detections}, the union between the sets $\mathit{S1}$ and $\mathit{S2}$ is being constructed. Element from one set is considered to be equal to an element from the other set when their IoU score is greater than or equal to the threshold $t$. Lastly, the combination step works because of the commutative law $A \cup B = B \cup A$ and the associative law $(A \cup B) \cup C = A \cup (B \cup C)$. The whole combination process eventually returns a merged list $L$, as mentioned earlier, which contains detections that have been fused, along with all detections that are unique to each robot.

\section{Experiments and Results}\label{experiments_and_results}
\subsection{Map 1: The Detection Test}\label{map_1}
As mentioned previously in Section \ref{simulation_layout}, Map 1 (Fig. \ref{simulation_testbeds}) has been used to assess the performance of the whole intruder detection pipeline along with the fusion of accumulated detections performed by our central bot MIDNet. During each trial, the security robots execute their random patrolling algorithm, while the mobile intruder robots are given random goal points by our test script to wander around Map 1. From every trial, we have collected data consisting of the number of frames considered for the test, the total number of detections processed by our testing script, the numbers of true positives ($\mathit{tp}$), false positives ($\mathit{fp}$) and false negatives ($\mathit{fn}$) that have occurred during the trial. The number of true negatives ($\mathit{tn}$) have been ignored for two reasons: one being that they are not necessary to perform calculations that show how well our system is doing, and the other being that our detection algorithm does not involve tiling, or in other words, the sliding window approach. In our test, $\mathit{tp}$ is the number of bounding boxes for detections that intersect with the ground truth bounding boxes of the intruders. $\mathit{fp}$ is the total number of bounding boxes minus $\mathit{tp}$, or in other words, the number of detection boxes that do not intersect with any ground truth box. Finally, $\mathit{fn}$ is the number of times the system has failed to identify intruders when they are in laser scan range of at least one security robot, which in our case is $2.5$ metres. 

Using $\mathit{tp}$, $\mathit{fp}$ and $\mathit{fn}$, we have calculated $\mathit{precision}$, $\mathit{recall}$ and $\mathit{f1\, score}$. $\mathit{precision}$ (defined in Eq. (\ref{precision})) represents how accurately MIDNet has detected intruders given that it has submitted at least one detection to our test script. $\mathit{recall}$ (defined in Eq. (\ref{recall})) represents the accuracy of MIDNet in detecting intruders given that at least one intruder is in laser scan range. Finally, the $\mathit{f1\, score}$ (defined in Eq. (\ref{f1_score})) gives the weighted average of $\mathit{precision}$ and $\mathit{recall}$, or in other words, the overall performance of our intruder detection pipeline.

We have run one test trial involving one security robot against 8 intruder entities. We are referring to this test trial as ``1-robot". The aim of this trial is to assess the proficiency of MIDNet at detecting multiple intruders with a single robot. The intruders can be of various shapes and sizes and can either be mobile or stationary. 1999 frames have been observed for this test, which took approximately 23 minutes of real time equivalent to 5 minutes of simulation time. We have run another test trial ``5-robot", which involves 5 security robots against the same set of intruders, this time assessing the capability of MIDNet at detecting intruders with multiple robots. Similar to the previous test, 1999 frames have been observed for this test as well and it took approximately 1 hour and 21 minutes, which is equal to about 8 minutes of simulation time. The results are summarized in Table \ref{detection_scores}.

\begin{equation}\label{precision}
	\mathit{precision} = \frac{\mathit{tp}}{\mathit{tp} + \mathit{fp}}
\end{equation}

\begin{equation}\label{recall}
	\mathit{recall} = \frac{\mathit{tp}}{\mathit{tp} + \mathit{fn}}
\end{equation}

\begin{equation}\label{f1_score}
	\mathit{f1\, score} = 2 \left(\frac{\mathit{precision} \times \mathit{recall}}{\mathit{precision} + \mathit{recall}}\right)
\end{equation} 

\begin{table}[h]
	\caption{Detection scores involving 8 intruders, 3 of which are stationary and much larger than the intruder robots.}
	\label{detection_scores}
			\begin{tabular}{|c||c|c|c|c|c|c|}
				\hline
				trial & $\mathit{tp}$ & $\mathit{fp}$ & $\mathit{fn}$ & $\mathit{precision}$ & $\mathit{recall}$ & $\mathit{f1\, score}$\\
				\hhline{|=||=|=|=|=|=|=|}
				1-robot & 6967 & 176 & 1889 & 0.975 & 0.787 & 0.871\\
				\hline
				5-robot & 14953 & 1047 & 619 & 0.935 & 0.96 & 0.947\\
				\hline
				\multicolumn{7}{|c|}{Frames observed: 1999}\\
				\hline
			\end{tabular}
\end{table}

Looking at Table \ref{detection_scores}, it is observed that when 1 security robot is included in MIDNet, the Precision value is slightly higher than that when 5 robots are included in MIDNet. As for the value of Recall, it is a lot less for ``1-robot" case compared to that of the ``5-robot" case. Thus, the resulting F1 Score for the ``1-robot" configuration is lower than that of the ``5-robot" configuration by a relatively small margin. Our observations as to why the scores differ are as described in the rest of this subsection.

The laser distance sensor that we have used provides data at a rate of 5 Hz, which produces delay in the case of objects in motion or stationary objects at the edge of the laser scan range when the security robot is in motion. This delay produces a false positive when MIDNet has detected an intruder, but that intruder has exited the laser scan boundaries of all robots or has moved farther from its last registered location. This delay also produces a false negative when an intruder enters the laser scan range of at least one robot and MIDNet fails to register that intruder instantly. However, the laser-scan delay is not expected to be significant unless the intruder travels at a fairly high speed.

Performance of MIDNet also degrades due to partial vision that is caused by different viewpoints of the security robots, as well as when the laser scans are blocked partially or completely by an object. In both the single-robot and multi-robot scenario, when one intruder lines up in front of a larger intruder in such a way that the larger intruder seems to be two intruders from the viewpoint of a security robot, a false positive occurs. In the multi-robot scenario, the number of false positives further increase when security robots having different viewpoints fail to completely draw the intruder entity, interpreting the same intruder as multiple intruders. This explains why the number of false positives for the ``5-robot" configuration in Table \ref{detection_scores} is significantly higher than that of the ``1-robot" configuration. On the contrary, the false negative score for the ``5-robot" configuration is significantly less than that of the ``1-robot" configuration. In this case, the false negative in the ``1-robot" configuration occurs when an intruder is in laser scan range but lines up behind a larger object or intruder in such a way that the security robot fails to detect that intruder. This issue is insignificant in the ``5-robot" configuration, as any other security robot having a different viewpoint can detect the intruder when an ally fails to find it, consequently cancelling out the ``blind spot" of MIDNet.

\subsection{Map 2: The Labyrinth Test}\label{map_2}
\begin{figure}
	\centering
	\includegraphics[width=\linewidth]{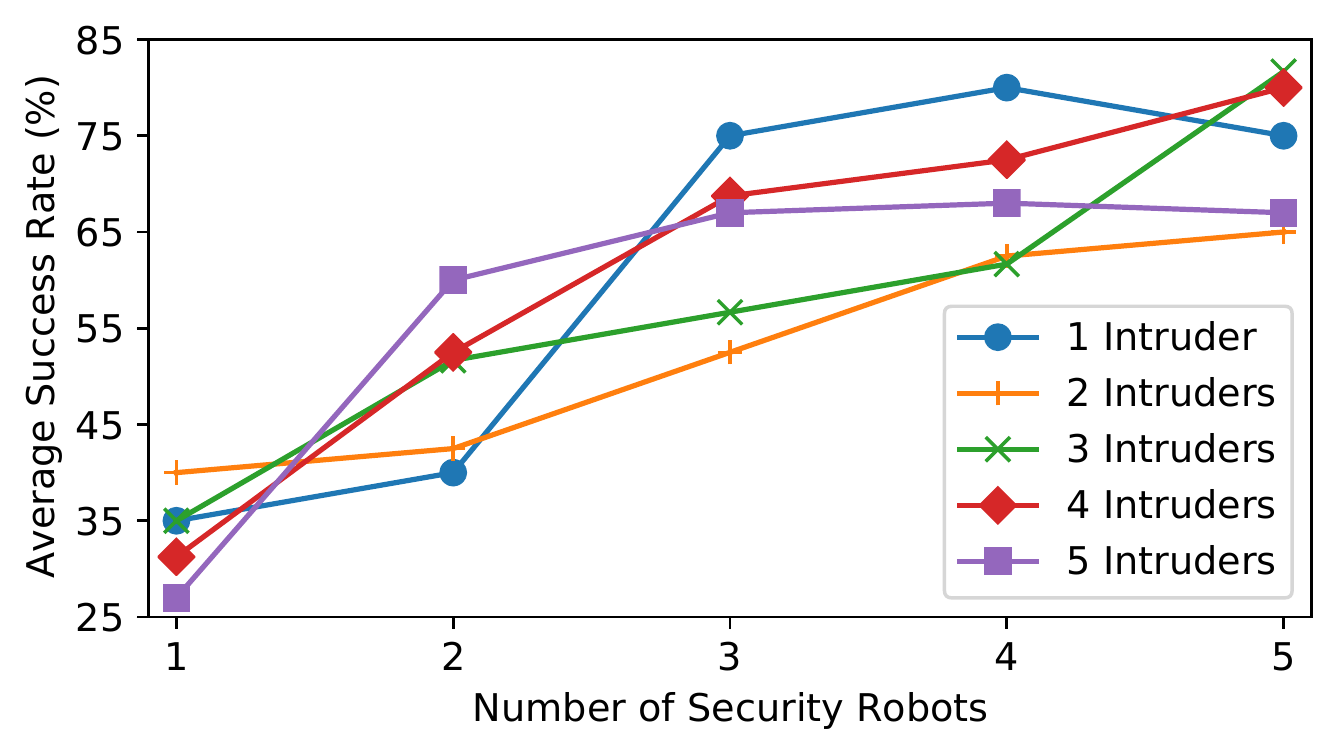}
	\caption{Average Success Rates over 20 trials. Relevant to Section \ref{map_2}.}
	\label{average_success_rate_graph}
\end{figure}

Referring to the previous Section \ref{simulation_layout}, Map 2 (Fig. \ref{simulation_testbeds}) has been used to analyze average success rates of convergence to the intruders for different team sizes of security robots. We have run 20 trials for each combination from a set of all $25$ possible combinations of 5 intruders and 5 security robots (500 trials in total). The success rate for each trial is calculated using Eq. (\ref{success_rate}):
\begin{equation}\label{success_rate}
	\mathit{success\, rate} = \frac{\mathit{number\, of\, intruders\, caught}}{\mathit{total\, number\, of\, intruders}} \times 100
\end{equation}
For every trial, our test script spawns intruder robots at different locations and assigns random escape locations to these robots. The starting and escape points of each intruder robot are at least $12m$ apart. The security robots are spawned at the center block of Map 2. MIDNet generates random patrol coordinates for each security robot, and the coordinates are regenerated if they fail to meet the following set of conditions:

\begin{itemize}
	\item The coordinates are at least $z$ meters away from the robot's current position. $z$ is calculated using Eq. (\ref{safe_distance}).
	\begin{equation}\label{safe_distance}
		z = \frac{p}{2q}
	\end{equation}
	where $p$ is the length of diagonal of enclosed area in meters and $q$ is the number of security robots.
	\item The coordinates are at least $z$ meters away from the current patrol coordinates of the other security robots, where $z$ is defined in the aforementioned Eq. (\ref{safe_distance}).
	\item The coordinates are at least $0.4m$ away from any occupied grid cell. 
\end{itemize}
An average value of success rate over the 20 trials is calculated for each combination of intruders and security robots. The results are summarized in Fig. \ref{average_success_rate_graph}. It took about 92 hours of real time (almost 4 days) to generate the data, which is equivalent to approximately 17 hours of simulated time in our experiments.

The usable laser scan range of each robot is $2.5m$, which gives a scan area of $6.25m^2$. The dimensions of Map 2 (given in Fig. \ref{simulation_testbeds}(b)) are $20m \times 20m$, which is an area of $400m^2$. This means that each robot has to monitor a map that has an area 64 times greater than its laser scan area. Therefore, adding more robots to the system should ensure greater coverage, which consequently increases the probability of detecting intruders in the map.

It is observed that for all the cases represented by the line graphs, the average success rates generally increase when the number of security robots is increased. However, for both the 1-intruder and the 5-intruder cases, the average success rates climb upwards when the number of security robots is increased from 1 to 4, followed by a very small decrease when the team size is increased from 4 to 5. The decrease may have been caused due to the randomization of patrolling coordinates by MIDNet. Hence, this does not affect the overall observation.

\thanks{{Demo Link: \url{https://youtu.be/dK1jyAjvtEs}}}%

\section{Conclusion}\label{conclusion}
In this paper, we have proposed a centralized algorithm MIDNet that can detect intruders inside a warehouse in absence of human staff, using ROS and 2D-LiDAR information from a team of robots. Referring to Table \ref{detection_scores}, experimental results show that MIDNet performs well in detecting intruders for both the single-robot case and the multi-robot case with enhanced results in the latter case. Fig. \ref{average_success_rate_graph} shows that increasing the number of robots under MIDNet enhances the security of the guarded region. Lastly, we can virtually add an infinite number of robots under MIDNet, and this makes our proposed system robust and scalable.

\bibliographystyle{IEEEtran}
\bibliography{IEEEabvr, references}

\begin{thebibliography}{10}
\providecommand{\url}[1]{#1}
\csname url@rmstyle\endcsname
\providecommand{\newblock}{\relax}
\providecommand{\bibinfo}[2]{#2}
\providecommand\BIBentrySTDinterwordspacing{\spaceskip=0pt\relax}
\providecommand\BIBentryALTinterwordstretchfactor{4}
\providecommand\BIBentryALTinterwordspacing{\spaceskip=\fontdimen2\font plus
\BIBentryALTinterwordstretchfactor\fontdimen3\font minus
  \fontdimen4\font\relax}
\providecommand\BIBforeignlanguage[2]{{%
\expandafter\ifx\csname l@#1\endcsname\relax
\typeout{** WARNING: IEEEtran.bst: No hyphenation pattern has been}%
\typeout{** loaded for the language `#1'. Using the pattern for}%
\typeout{** the default language instead.}%
\else
\language=\csname l@#1\endcsname
\fi
#2}}

\bibitem{carder2018}
\BIBentryALTinterwordspacing
H.~{Carder}, ``Global outlook for mobile security robots,'' Report, 2018.
  [Online]. Available:
  \url{https://www.roboticsbusinessreview.com/wp-content/uploads/2018/03/RBR-UGVs-For-Security.pdf}
\BIBentrySTDinterwordspacing

\bibitem{cong2020}
Y.~Cong, C.~Chen, J.~Li, W.~Wu, S.~Li, and B.~Yang, ``Mapping without dynamic:
  Robust lidar-slam for ugv mobile mapping in dynamic environments,'' \emph{The
  International Archives of Photogrammetry, Remote Sensing and Spatial
  Information Sciences}, vol.~43, pp. 515--520, 2020.

\bibitem{lenac2017}
K.~Lenac, A.~Kitanov, R.~Cupec, and I.~Petrovi{\'c}, ``Fast planar surface 3d
  slam using lidar,'' \emph{Robotics and Autonomous Systems}, vol.~92, pp.
  197--220, 2017.

\bibitem{catapang2016}
A.~N. Catapang and M.~Ramos, ``Obstacle detection using a 2d lidar system for
  an autonomous vehicle,'' in \emph{2016 6th IEEE International Conference on
  Control System, Computing and Engineering (ICCSCE)}.\hskip 1em plus 0.5em
  minus 0.4em\relax IEEE, 2016, pp. 441--445.

\bibitem{peng2015}
Y.~Peng, D.~Qu, Y.~Zhong, S.~Xie, J.~Luo, and J.~Gu, ``The obstacle detection
  and obstacle avoidance algorithm based on 2-d lidar,'' in \emph{2015 IEEE
  international conference on information and automation}.\hskip 1em plus 0.5em
  minus 0.4em\relax IEEE, 2015, pp. 1648--1653.

\bibitem{milford2008}
M.~J. Milford and G.~F. Wyeth, ``Mapping a suburb with a single camera using a
  biologically inspired slam system,'' \emph{IEEE Transactions on Robotics},
  vol.~24, no.~5, pp. 1038--1053, 2008.

\bibitem{milford2011}
M.~J. Milford, F.~Schill, P.~Corke, R.~Mahony, and G.~Wyeth, ``Aerial slam with
  a single camera using visual expectation,'' in \emph{2011 IEEE international
  conference on robotics and automation}.\hskip 1em plus 0.5em minus
  0.4em\relax IEEE, 2011, pp. 2506--2512.

\bibitem{chia2009}
C.-C. Chia, W.-K. Chan, and S.-Y. Chien, ``Cooperative surveillance system with
  fixed camera object localization and mobile robot target tracking,'' in
  \emph{Pacific-Rim Symposium on Image and Video Technology}.\hskip 1em plus
  0.5em minus 0.4em\relax Springer, 2009, pp. 886--897.

\bibitem{liu2005}
J.~N. Liu, M.~Wang, and B.~Feng, ``ibotguard: an internet-based intelligent
  robot security system using invariant face recognition against intruder,''
  \emph{IEEE Transactions on Systems, Man, and Cybernetics, Part C
  (Applications and Reviews)}, vol.~35, no.~1, pp. 97--105, 2005.

\bibitem{pierzchala2018}
M.~Pierzcha{\l}a, P.~Gigu{\`e}re, and R.~Astrup, ``Mapping forests using an
  unmanned ground vehicle with 3d lidar and graph-slam,'' \emph{Computers and
  Electronics in Agriculture}, vol. 145, pp. 217--225, 2018.

\bibitem{shi2019}
C.~Shi, K.~Huang, Q.~Yu, J.~Xiao, H.~Lu, and C.~Xie, ``Extrinsic calibration
  and odometry for camera-lidar systems,'' \emph{IEEE Access}, vol.~7, pp.
  120\,106--120\,116, 2019.

\bibitem{asvadi2017}
A.~Asvadi, L.~Garrote, C.~Premebida, P.~Peixoto, and U.~J. Nunes, ``Depthcn:
  vehicle detection using 3d-lidar and convnet,'' in \emph{2017 IEEE 20th
  International Conference on Intelligent Transportation Systems (ITSC)}.\hskip
  1em plus 0.5em minus 0.4em\relax IEEE, 2017, pp. 1--6.

\bibitem{roth2019}
M.~Roth, D.~Jargot, and D.~M. Gavrila, ``Deep end-to-end 3d person detection
  from camera and lidar,'' in \emph{2019 IEEE Intelligent Transportation
  Systems Conference (ITSC)}.\hskip 1em plus 0.5em minus 0.4em\relax IEEE,
  2019, pp. 521--527.

\bibitem{guerrero2019}
{\'A}.~M. Guerrero-Higueras, C.~{\'A}lvarez-Aparicio, M.~C. Calvo~Olivera,
  F.~J. Rodr{\'\i}guez-Lera, C.~Fern{\'a}ndez-Llamas, F.~M. Rico, and
  V.~Matell{\'a}n, ``Tracking people in a mobile robot from 2d lidar scans
  using full convolutional neural networks for security in cluttered
  environments,'' \emph{Frontiers in neurorobotics}, vol.~12, p.~85, 2019.

\bibitem{sankar2019}
S.~Sankar and C.-Y. Tsai, ``Ros-based human detection and tracking from a
  wireless controlled mobile robot using kinect,'' \emph{Applied System
  Innovation}, vol.~2, no.~1, p.~5, 2019.

\bibitem{dharmasena2019}
T.~Dharmasena and P.~Abeygunawardhana, ``Design and implementation of an
  autonomous indoor surveillance robot based on raspberry pi,'' in \emph{2019
  International Conference on Advancements in Computing (ICAC)}.\hskip 1em plus
  0.5em minus 0.4em\relax IEEE, 2019, pp. 244--248.

\bibitem{khamis2015}
A.~Khamis, A.~Hussein, and A.~Elmogy, ``Multi-robot task allocation: A review
  of the state-of-the-art,'' in \emph{Cooperative Robots and Sensor Networks
  2015}.\hskip 1em plus 0.5em minus 0.4em\relax Springer, 2015, pp. 31--51.

\bibitem{karapetyan2017}
N.~Karapetyan, K.~Benson, C.~McKinney, P.~Taslakian, and I.~Rekleitis,
  ``Efficient multi-robot coverage of a known environment,'' in \emph{2017
  IEEE/RSJ International Conference on Intelligent Robots and Systems
  (IROS)}.\hskip 1em plus 0.5em minus 0.4em\relax IEEE, 2017, pp. 1846--1852.

\bibitem{folgado2007}
E.~Folgado, M.~Rinc{\'o}n, J.~R. {\'A}lvarez, and J.~Mira, ``A multi-robot
  surveillance system simulated in gazebo,'' in \emph{International
  Work-Conference on the Interplay Between Natural and Artificial
  Computation}.\hskip 1em plus 0.5em minus 0.4em\relax Springer, 2007, pp.
  202--211.

\bibitem{trigui2012}
S.~Trigui, A.~Koub{\^a}a, M.~B. Jam{\^a}a, I.~Ch{\^a}ari, and K.~Al-Shalfan,
  ``Coordination in a multi-robot surveillance application using wireless
  sensor networks,'' in \emph{2012 16th IEEE Mediterranean Electrotechnical
  Conference}.\hskip 1em plus 0.5em minus 0.4em\relax IEEE, 2012, pp. 989--992.

\bibitem{tuna2012}
G.~Tuna, C.~Tasdemir, K.~Gulez, T.~V. Mumcu, and V.~C. Gungor, ``Autonomous
  intruder detection system using wireless networked mobile robots,'' in
  \emph{2012 IEEE Symposium on Computers and Communications (ISCC)}.\hskip 1em
  plus 0.5em minus 0.4em\relax IEEE, 2012, pp. 000\,001--000\,005.

\bibitem{grisetti2005}
G.~{Grisetti}, C.~{Stachniss}, and W.~{Burgard}, ``Improving grid-based slam
  with rao-blackwellized particle filters by adaptive proposals and selective
  resampling,'' in \emph{Proceedings of the 2005 IEEE International Conference
  on Robotics and Automation}, 2005, pp. 2432--2437.

\bibitem{grisetti2007}
------, ``Improved techniques for grid mapping with rao-blackwellized particle
  filters,'' \emph{IEEE Transactions on Robotics}, vol.~23, no.~1, pp. 34--46,
  2007.

\bibitem{moore2016}
T.~Moore and D.~Stouch, ``A generalized extended kalman filter implementation
  for the robot operating system,'' in \emph{Intelligent autonomous systems
  13}.\hskip 1em plus 0.5em minus 0.4em\relax Springer, 2016, pp. 335--348.

\bibitem{julier1997}
S.~J. Julier and J.~K. Uhlmann, ``New extension of the kalman filter to
  nonlinear systems,'' in \emph{Signal processing, sensor fusion, and target
  recognition VI}, vol. 3068.\hskip 1em plus 0.5em minus 0.4em\relax
  International Society for Optics and Photonics, 1997, pp. 182--193.

\bibitem{horner2016}
\BIBentryALTinterwordspacing
J.~{Hörner}, ``Map-merging for multi-robot system,'' Bachelor's thesis,
  Charles University in Prague, Faculty of Mathematics and Physics, Prague,
  2016. [Online]. Available:
  \url{https://is.cuni.cz/webapps/zzp/detail/174125/}
\BIBentrySTDinterwordspacing

\bibitem{dijkstra1959}
E.~W. {Dijkstra}, ``A note on two problems in connexion with graphs,''
  \emph{Numerische Mathematik}, vol.~1, no.~1, pp. 269--271, 1959.

\bibitem{lu2014}
D.~V. {Lu}, D.~{Hershberger}, and W.~D. {Smart}, ``Layered costmaps for
  context-sensitive navigation,'' in \emph{2014 IEEE/RSJ International
  Conference on Intelligent Robots and Systems}, 2014, pp. 709--715.

\bibitem{fox1997}
D.~{Fox}, W.~{Burgard}, and S.~{Thrun}, ``The dynamic window approach to
  collision avoidance,'' \emph{IEEE Robotics Automation Magazine}, vol.~4,
  no.~1, pp. 23--33, 1997.

\bibitem{kelly1994}
A.~{Kelly}, ``An intelligent predictive controller for autonomous vehicles,''
  Carnegie Mellon University, Pittsburgh, PA, Tech. Rep. CMU-RI-TR-94-20, May
  1994.

\bibitem{gerkey2008}
B.~P. {Gerkey} and K.~{Konolige}, ``Planning and control in unstructured
  terrain,'' in \emph{ICRA Workshop on Path Planning on Costmaps}, 2008.

\bibitem{suzuki1985}
S.~{Suzuki} and K.~{Abe}, ``Topological structural analysis of digitized binary
  images by border following,'' \emph{Computer Vision, Graphics, and Image
  Processing}, vol.~30, no.~1, pp. 32 -- 46, 1985.

\end{thebibliography}
\end{document}